\documentclass{article}

\usepackage[preprint]{neurips_2023}

\usepackage[utf8]{inputenc} 
\usepackage[T1]{fontenc}    
\usepackage{hyperref}       
\usepackage{url}            
\usepackage{booktabs}       
\usepackage{amsfonts}       
\usepackage{nicefrac}       
\usepackage{microtype}      
\usepackage{xcolor}         

\usepackage{amsmath}
\usepackage{graphicx}
\usepackage{float}
\usepackage{subcaption}
\usepackage{cite}

\hypersetup{colorlinks, linkcolor=blue,
citecolor=blue, urlcolor=blue}

\title{PinnDE: Physics-Informed Neural Networks for Solving Differential Equations}

\author{%
  Jason Matthews\\
  Department of Mathematics and Statistics\\
  Memorial University of Newfoundland\\
  St. John’s, NL, A1C 5S7, Canada\\
  \texttt{jbmatthews@mun.ca} \\
  \And
  Alex Bihlo\\
  Department of Mathematics and Statistics\\
  Memorial University of Newfoundland\\
  St. John’s, NL, A1C 5S7, Canada\\
  \texttt{abihlo@mun.ca} \\
}

\begin{document}

\maketitle

\textbf{Keywords:} PinnDE, differential equations, physics-informed neural networks, deep operator networks

\begin{abstract}
  In recent years the study of deep learning for solving differential equations has grown substantially. The use of physics-informed neural networks (PINNs) and deep operator networks (DeepONets) have emerged as two of the most useful approaches in approximating differential equation solutions using machine learning. Here, we introduce PinnDE, an open-source Python library for solving differential equations with both PINNs and DeepONets. We give a brief review of both PINNs and DeepONets, introduce PinnDE along with the structure and usage of the package, and present worked examples to show PinnDE's effectiveness in approximating solutions of systems of differential equations with both PINNs and DeepONets.
\end{abstract}

\section{Introduction}

Powerful numerical algorithms for solving differential equations have been developed for well over one hundred years \cite{rungekutta}. Modern deep learning has been applied and studied through many diverse fields \cite{chowdhary2020natural, voulodimos2018deep}, however the use of deep learning for solving differential equations is a relatively new field. While the first ideas of these methods were proposed in \cite{lagaris1998artificial, lagaris2000neural}, in \cite{raissi2019physics} the notion of a physics-informed neural network (PINN) was popularized. Compared to finite difference methods, finite element methods or finite volume methods that use numerical differentiation for computing derivatives, using automatic differentiation \cite{baydin2018automatic} in deep learning-based approaches provides a meshless alternative to solving differential equations. While \cite{raissi2019physics} first proposed this for solving forward and inverse problems, many other variations of differential equations have been demonstrated to be able to be solved by PINNs. PINNs can not only be used for solving differential equations directly, the so-called \textit{forward problem}, they can also be used for parameter identification or learning of differential equations itself from data, which is referred to as the \textit{inverse problem}, as discussed in \cite{lu2021physics, raissi2020hidden, yu2022gradient}. Solving fractional differential equations has been proposed and improved upon in the works of \cite{pang2019fpinns, sm2024novel}. Integro-differential equations have been shown to be amenable to PINN-based methods \cite{yuan2022pinn} as well. Stochastic differential equations were tackled with only slight variations to standard PINN architecture \cite{yang2020physics, guo2022normalizing, zhang2019quantifying}. There has also been some initial work indicating that including geometric properties of differential equations can improve the numerical solutions obtainable using physics-informed machine learning, cf. \cite{arora2024invariant,cardoso2023exactly}.

While there has been much research done on the different variations of differential equations solvable by PINNs, work on improving the ability of PINNs in general has been a large focus over the past several years as well. Adaptive methods for collocation point sampling have been a common field of study, where residual-based adaptive refinement (RAR) is a popular method which has been proposed and improved upon in the works of \cite{lu2019deeponet, zeng2022adaptive, nabian2021efficient, hanna2022residual, wu2023comprehensive}. More specific adaptive collocation point have been used in \cite{wight2020solving, tang2022adaptive}. Techniques for providing adaptive scaling parameters on the PINN architecture are developed in \cite{wang2021understanding} and \cite{mcclenny2023self}. Meta learning, a method for a neural network learned optimizer to optimize a PINN as opposed to a hand-crafted optimizer, has been a relatively new area of research to improve physics-informed machine learning for solving differential equations \cite{psaros2022meta, bihlo2024improving, liu2022novel}. 

While PINNs are the main focus using deep learning for solving differential equations, a second, more general approach has grown in popularity as a tool for solving differential equations as well. Deep operator networks (DeepONets), first proposed in \cite{lu2019deeponet}, provide a general method for solving differential equations. A key distinction separates PINNs from DeepONets. While a PINN is trained to approximate the solution function, \(u(x)\), for a single instance of a differential equation, a DeepONet is designed to learn the underlying solution operator, \(\mathsf{G}\). The goal is to approximate the mapping from an input function \(f\) (such as the initial condition, boundary conditions, or forcing term) to the corresponding solution function, such that \(u(x)\approx \mathsf{G}(f)(x)\). This operator learning framework provides a significant advantage: Once trained, the DeepONet can be evaluated on new, unseen input functions without requiring retraining, making it highly efficient for solving parametric problems. Furthermore, the learned operator can be applied sequentially for long time domain predictions. By using the predicted solution at one time step as the new initial condition for the subsequent step, the network can perform time-stepping simulations, as demonstrated by  \cite{wang2023long}. Many different DeepONet variations have been proposed throughout \cite{venturi2023svd, hadorn2022shift, seidman2022nomad, lee2023hyperdeeponet,zeudong2025low} which show progress in the improvement in this architecture.

The Python machine learning frameworks TensorFlow, PyTorch, and JAX \cite{tensorflow2015-whitepaper, pytorch, jax2018github} are commonly used for implementing deep learning algorithms and architecture. Despite this and deep learning for differential equations becoming popular research fields, there exist few software packages which implement these ideas using low-code solutions. While existing packages like DeepXDE \cite{lu2021deepxde} provide powerful and flexible frameworks, they often demand significant software development experience. This results in researchers and educators having to re-create already well-understood implementations from scratch. Here, we introduce PinnDE, an open-source Python library for solving differential equations with PINNs and DeepONets implemented in TensorFlow and JAX. PinnDE provides a user-friendly interface in which only a minimal amount of functions is given to a user. PinnDE also provides simple to read implementations which we believe gives this package the ability to be used in education and research effectively, requiring minimal background knowledge on physics-informed machine learning by its users.

This paper is organized as follows: We briefly discuss the theory behind PINNs and DeepONets in Section \ref{2}, and present a short but self-contained general overview of their implementation. Next, in Section \ref{3} we introduce PinnDE, providing a summary of the structure of the package with explanations on its usage. Then, in Section \ref{4} we provide some worked examples of commonly solved ordinary and partial differential equations using different methods to show the effectiveness of PinnDE. Finally, Section \ref{5} contains the conclusions and some discussions about further plans for developing PinnDE.

\section{Background}\label{2}

In this section we give a short review of a physics-informed neural network (PINN), deep operator network (DeepONet), and inverse PINNs which are the basis of the algorithms and models implemented in PinnDE. We then give an overview of current adaptive point sampling strategies and optimization routines which PinnDE covers.

\subsection{Physics-informed neural networks for forward problem}\label{2.1}

Physics-informed neural networks (PINNs) were first introduced in \cite{lagaris1998artificial} and were later popularized when re-introduced in \cite{raissi2019physics}. The general idea consists of taking a deep neural network as a surrogate approximation for the solution of a system of differential equations. This has the advantage that derivatives of the neural network approximation to the solution of the system of differential equations can be computed with automatic differentiation \cite{baydin2018automatic} rather than relying on numerical differentiation as is typically the case for standard numerical methods such as finite difference, finite volume or finite element methods. As a consequence, physics-informed neural networks are truly meshless methods as the derivative computations can be done in single points, without relying on the introduction of a computational mesh as is the case for most standard numerical methods. The neural network surrogate solution is being obtained by solving an optimization problem that involves fitting the weights and biases of the network. This is done such that at a collection of finitely many (typically randomly chosen) collocation points a loss function combining the differential equation along with any supplied initial and/or boundary conditions is minimized. We make this general procedure more precise in the following.

We consider a general initial--boundary value problem for a system of \(L\) partial differential equations over a spatio-temporal domain \([t_{\rm 0}, t_{\rm f}] \times \Omega\), where \(\Omega \subset \mathbb{R}^{d} \), \(t_{\rm 0}\in\mathbb{R}\) denotes the initial integration time, and \(t_{\rm f}\in\mathbb{R}\) denotes the final integration time, given by
\begin{subequations}
\begin{align}
    &\Delta^{l}(t, x, u_{(n)}) = 0,\quad l = 1,\dots,L,\quad t\in[t_{\rm 0}, t_{\rm f}],\, x\in\Omega,& \label{eqn_1a} \\
    & \mathsf{I}^{l_{\rm i}}(x, u_{(n_{\rm i})}|_{t=t_{\rm 0}}) = 0,\quad l_{\rm i} = 1,\dots,L_{\rm i},\quad x\in\Omega,& \label{eqn_1b} \\
    & \mathsf{B}^{l_{\rm b}}(t, x, u_{(n_{\rm b})}) = 0,\quad l_{\rm b} = 1,\dots,L_{\rm b},\quad t\in[t_{\rm 0}, t_{\rm f}],\, x\in\partial\Omega, \label{eqn_1c}&
\end{align}\label{eqn_1}
\end{subequations}
where \(t\) denotes the time variable and \(x = (x_{1},\dots,x_{d})\) denotes the tuple of spatial independent variables. The dependent variables are denoted by \(u = (u^{1},\dots,u^{q})\), and \(u_{(n)}\) denotes the tuple of all derivatives of the dependent variable with respect to both \(t\) and \(x\) up to order \(n\). The initial conditions are represented through the initial value operator \(\mathsf{I} = (\mathsf{I}^{1},\dots,\mathsf{I}^{L_{\rm i}})\), and similarly the boundary conditions are included using the boundary value operator \(\mathsf{B} = (\mathsf{B}^{1},\dots,\mathsf{B}^{L_{\rm b}})\).

We denote a deep neural network as \(\mathcal{N}^\theta\) with parameters \(\theta\), which includes all the weights and biases of all layers of the neural network. The goal of a physics-informed neural network is to learn to interpolate the global solution of the system of differential equations over \([t_{\rm 0},t_{\rm f}]\times\Omega\) as the parameterization \(u^\theta(t,x) = \mathcal{N}^{\theta}(t, x)\), where \(u^{\theta}(t, x) \approx u(t, x)\). This is done by minimizing the loss function
\begin{equation}
    \label{eqn_2}
    \mathcal{L}(\theta) = \mathcal{L}_{\Delta}(\theta) + \gamma_{\rm i}\mathcal{L}_{\rm i}(\theta) + \gamma_{\rm b}\mathcal{L}_{\rm b}(\theta),
\end{equation}
where \(\gamma_{\rm i}, \gamma_{\rm b} \in\mathbb{R}^+\) are weighting parameters for the individual loss contributions, and we composite the differential equation, initial value, and boundary value loss, respectively, which are given by
\begin{subequations}
\begin{align}
    & \mathcal{L}_{\Delta}(\theta) = \frac{1}{N_{\Delta}} \sum_{i=1}^{N_{\Delta}}\sum_{l=1}^{L} | \Delta^{l}(t_{\Delta}^{i}, x_{\Delta}^{i}, u_{(n)}^{\theta}(t_{\Delta}^{i}, x_{\Delta}^{i}))|^2,\label{eqn_3a}&\\
    & \mathcal{L}_{\rm i}(\theta) = \frac{1}{N_{\rm i}} \sum_{i=1}^{N_{\rm i}}\sum_{l_{\rm i}=1}^{L_{\rm i}} |\mathsf{I}^{l_{\rm i}}(t_{\rm i}^{i}, x_{\rm i}^{i}, u_{(n_{\rm i})}^{\theta}(t_{\rm i}^{i}, x_{\rm i}^{i}))|^2, \label{eqn_3b}&\\
    & \mathcal{L}_{\rm b}(\theta) = \frac{1}{N_{\rm b}} \sum_{i=1}^{N_{\rm b}}\sum_{l_{\rm b}=1}^{L_{\rm b}} |\mathsf{B}^{l_b}(t_{\rm b}^{i}, x_{\rm b}^{i}, u_{(n_{\rm b})}^{\theta}(t_{\rm b}^{i}, x_{\rm b}^{i}))|^2, \label{eqn_3c}&
\end{align}
\end{subequations}
where \(\mathcal{L}_{\Delta}(\theta)\) corresponds to the differential equation loss based on Eqn.~\eqref{eqn_1a}, \(\mathcal{L}_{\rm i}(\theta)\) is the initial value loss stemming from Eqn.~\eqref{eqn_1b}, and \(\mathcal{L}_{\rm b}(\theta)\) denotes the boundary value loss derived from Eqn.~\eqref{eqn_1c}. The neural network evaluates the losses over a set of collocation points, where we have \(\{(t_{\Delta}^{i}, x_{\Delta}^{i})\}_{i=1}^{N_{\Delta}}\) for the system, \(\{(t_{\rm i}^{i}, x_{\rm i}^{i})\}_{i=1}^{N_{\rm i}}\) for the initial values, and \(\{(t_{\rm b}^{i}, x_{\rm b}^{i})\}_{i=1}^{N_{\rm b}}\) for the boundary values, with \(N_{\Delta}\), \(N_{\rm i}\) and \(N_{\rm b}\) denoting the number of differential equation, initial condition and boundary condition collocation points, respectively.

This method is considered \textit{soft-constrained}, as the network is forced to learn the initial and boundary conditions  and composite their loss, which is described in \cite{raissi2019physics}. A downside of this approach is that while the initial and boundary values are known exactly, they will in general not be satisfied exactly by the learned neural network owing to being enforced only through minimizing the loss function. An alternative strategy is to \textit{hard-constrain} the network with the initial and/or boundary conditions, following what was first proposed in \cite{lagaris1998artificial}, which structures the neural network itself in a way so that the output of the network automatically satisfies the initial and/or boundary conditions. As such, only the differential equation loss then has to be minimized and the initial and boundary loss components in \eqref{eqn_2} are not required. The general idea of hard-constraining the initial and boundary conditions is hence to design a suitable ansatz for the neural network surrogate solution that makes sure that the initial and boundary conditions are exactly satisfied for all values of the trainable neural network. The precise form of the class of suitable hard constraints depends on the particular form of the initial and boundary value operators. We show some specific implementations in Section \ref{3.1.1}. A more formalized discussion can also be found in \cite{brecht2023improving}. For many problems, it has been shown that hard-constraining the initial and boundary conditions improves the training performance of neural network based differential equations solvers, and leads to lower overall errors in the obtained solutions of these solvers, cf.~\cite{brecht2023improving}. One notable exception are cases where the solution of a differential equation is not differentiable everywhere, in which case soft constraints typically outperform hard constraints, see e.g.~\cite{brecht2024physics}.

Above we have introduced physics-informed neural networks for systems of partial differential equations for general initial--boundary value problems. The same method can also be applied for ordinary differential equations. For ordinary differential equations, both initial value problems and boundary value problems can be considered. 

We first consider a system of \(L\) ordinary differential equations over the temporal domain \([t_{\rm 0}, t_{\rm f}]\),
\begin{equation}
    \Delta^{l}(t, u_{(n)}) = 0,\quad l = 1,\dots,L,\quad t\in[t_{\rm 0}, t_{\rm f}]\label{eqn_4}
\end{equation}
where \(u_{(n)}\) are the total derivatives with respect to \(t\), and \(\Delta^{l}\) are \((m+1)\)-th order differential functions, with initial conditions,
\begin{equation}
    \mathsf{I}^{l_{\rm i}}(u_{(n_{\rm i})}|_{t=t_{\rm 0}}) = 0,\quad l_{\rm i} = 1,\dots,L_{\rm i} \label{eqn_5}
\end{equation}
where the associated loss function for the physics-informed neural network
is represented as the composite loss of Eqn.~\eqref{eqn_3a} and Eqn.~\eqref{eqn_3b}, with \(\Delta^{l}\) and \(\mathsf{I}^{l_{\rm i}}\) corresponding to Eqn.~\eqref{eqn_4} and Eqn.~\eqref{eqn_5}.
One can also consider a boundary value problem with the boundary conditions
\begin{equation}
    \mathsf{B}^{l_{\rm b}}(t, u_{(n_{\rm b})}) = 0,\quad l_{\rm b} = 1,\dots,L_{\rm b},\quad t\in\{t_{\rm 0}, t_{\rm f}\}, \label{eqn_6}
\end{equation}
in which case the physics-informed loss function is represented as the composite loss of Eqn.~\eqref{eqn_3a} and Eqn.~\eqref{eqn_3c}, with \(\Delta^{l}\) and \(\mathsf{B}^{l_{\rm b}}\) corresponding to Eqn.~\eqref{eqn_4} and Eqn.~\eqref{eqn_6}.

Hard constraining these networks follows a similar procedure as described above, with the initial/boundary values being exactly enforced for all values of the neural network surrogate solution. We refer to Section \ref{3.1.1} for further details.

\subsection{Physics-informed neural networks for inverse problems} \label{2.2}

Physics-informed neural networks for inverse problems were introduced in \cite{raissi2019physics}, and resemble closely the structure outlined in Section \ref{2.1}, and the description of PINN methods in the previous Section \ref{2.1} holds true for inverse PINNs as well. The key distinction is for a deep neural network to learn to interpolate the global solution of a system of differential equations as well as parameters given in the system of differential equations. This is done by using experimental data to approximate said parameters.

In more detail, we consider the same system of partial differential equations described by Eqn.~\eqref{eqn_1}, with the addition of a parameter term
\begin{subequations}
\begin{align}
    & \Delta^{l}(t, x, u_{(n)}, \lambda) = 0,\quad l = 1,\dots,L,\quad t\in[t_{\rm 0}, t_{\rm f}],\, x\in\Omega, \label{eqn_7a}& \\
    & \mathsf{I}^{l_{\rm i}}(x, u_{(n_{\rm i})}, \lambda|_{t=t_{\rm 0}}) = 0,\quad l_{\rm i} = 1,\dots,L_{\rm i},\quad x\in\Omega, \label{eqn_7b}& \\
    & \mathsf{B}^{l_{\rm b}}(t, x, u_{(n_{\rm b})}, \lambda) = 0,\quad l_{\rm b} = 1,\dots,L_{\rm b},\quad t\in[t_{\rm 0}, t_{\rm f}],\, x\in\partial\Omega,& \label{eqn_7c} 
\end{align}
\label{eqn_7}
\end{subequations}
where \(\lambda = (\lambda_{1},\dots,\lambda_{p})\) denotes the tuple of parameters in this system of differential equations, with all other definitions given in Section \ref{2.1}. These parameters may also be functions of the independent variables, the dependent variables as well as the derivatives of the dependent variables.

In this case, for a neural network \(\mathcal{N}^\theta\) with weights and biases \(\theta\), the goal of an inverse PINN is to learn to interpolate the global solution of the system of differential equations over \([t_{\rm 0},t_{\rm f}]\times\Omega\) and identify the values of the parameters of the equations as the parameterization \((u^\theta, \lambda^\theta) = (\mathcal{N}^\theta_1,\dots,\mathcal{N}^\theta_{q+p}) = \mathcal{N}^{\theta}(t, x)\), where \(u^{\theta}(t, x) \approx u(t, x)\), and \(\lambda^\theta \approx  \lambda\).

This is done by adding a loss term for the experimental data to the loss function given by Eqn.~\eqref{2}. For a system of \(L\) partial derivative equations, we add the term \(\mathcal{L}_{\rm d}(\theta)\) given by
\begin{equation}
    \mathcal{L}_{\rm d}(\theta) = \frac{1}{N_{\rm d}} \sum_{i=1}^{N_{\rm d}}\sum_{k=1}^{q} |\mathcal{N}_k^\theta(t_{\rm d}^{i}, x_{\rm d}^{i}) - u^{k}(t_{\rm d}^{i}, x_{\rm d}^{i})|^2, \label{eqn_8}
\end{equation}
where the network is evaluated over the collected data points sampled at the spatio-temporal data collocation points \(\{(t_{\rm d}^{i}, x_{\rm d}^{i})\}_{i=1}^{N_{\rm d}}\), with \(N_{\rm d}\) denoting the number of data points sampled. For each of the solution components of the neural network approximation \(\mathcal{N}_k^{\theta}(t,x)\), \(k=1,\dots, q\), the mean squared error between this approximate solution and the experimental data of the underlying true solution \(u^k(t,x)\) sampled at the data point locations is computed. In this setting, \(\mathcal{N}^\theta\) learns to not only obtain a particular solution to a parameterized differential equation but to also fit the parameters \(\lambda\) to the experimental data samples.

\subsection{Deep Operator Networks} \label{2.3}

Deep operator networks (DeepONets) were first introduced in \cite{lu2019deeponet} based on theoretical results of \cite{chen1995universal}. This idea replaces learning of a particular solution of the system of differential equations with learning the solution operator acting on a specific initial--boundary condition for the system of differential equations~\eqref{eqn_1}. That is, let \(\mathsf{G}(u_0)(t,x)\) be a solution operator for \eqref{eqn_1} acting on the particular initial condition \(u_0\) yielding the solution of the initial--boundary value at the spatio-temporal point \((t,x)\), i.e.\ \(u(t,x) = \mathsf{G}(u_0)(t,x)\). That is, in comparison to standard physics-informed neural networks, deep operator networks require two inputs, the spatio-temporal point where the solution should be evaluated (this is the same input as for physics-informed neural networks), and the particular initial (or boundary) condition to which the solution operator should be applied. Since the initial condition is a function, it first has to be sampled on a finite dimensional subspace to be included into the neural network. The most prominent way to accomplish this is to sample a finite collection of \(N_{\rm s}\) \textit{sensor points} \(\{x_i\}_{i=1}^{N_{\rm s}},  x_i\in\Omega\) and evaluate the initial condition at those sensor points, yielding the tuple \((u_0(x_1), \dots, u_0(x_{N_{\rm s}}))\) that is used as an input for the deep operator network. A similar strategy is followed for pure boundary value problems. A deep operator network then uses two separate networks to combine the sampled initial conditions and independent variables. The \textit{branch network} processes the initial condition sampled values, and the \textit{trunk network} processes the independent variables. In the vanilla deep operator network the output vectors of the two sub-networks are then combined via a dot product. The output of deep operator network can be expressed as
\begin{equation*}
    \mathcal{G}^{\theta}(u_0(x))(t, x) = \mathcal{B}^{\theta}(u_0(x_1),\dots,u_0(x_{N_{\rm s}}))\cdot \mathcal{T}^{\theta}(t, x) 
\end{equation*}
With \(\mathcal{B}^{\theta}=(\mathcal{B}_1^{\theta},\dots,\mathcal{B}_p^{\theta})\) representing the output of the branch net and \(\mathcal{T}^{\theta}=(\mathcal{T}_1^{\theta},\dots,\mathcal{T}_p^{\theta})\) representing the output of the trunk net, and \(p\) being a hyper-parameter. The loss function associated with these networks is the same as \eqref{eqn_2} when soft-constrained. Analogously to physics-informed neural networks, also deep operator networks can be hard-constrained with either standard PINN hard constraining methods, or further modified methods designed for deep operator networks, cf.~\cite{brecht2023improving}. A trained deep operator network thus is a solution interpolant for the solution operator, i.e.\ \(\mathcal{G}^{\theta}\approx \mathsf{G}\).

These networks come with multiple main benefits. Firstly, these networks can easily replace one initial (and/or boundary) condition with a new initial (and/or boundary) condition and do not require re-training. Secondly, as presented in \cite{wang2023long}, since these networks learn the solution operator corresponding to an initial condition, they can readily be used for time-stepping for autonomous systems of differential equations and thus extend to long time intervals and maintain the ability to further predict the solution, unlike a standard PINN, which typically fails if \(t_{\rm f}\gg t_0\). A downside of DeepONets is that they are typically harder to train as PINNs, since not only a single particular solution has to be interpolated, but the entire solution operator. This also places additional burdens on the computational hardware required, making DeepONets more expensive to train than PINNs.

\subsection{Adaptive point sampling} \label{2.4}

The loss function described by Eqn.~\eqref{2} relies on sampled collocation points across the domain of the system of PDEs. This is most commonly done using a random or quasi-random sample of the domain. One of the most frequently used sampling strategies is Latin hypercube sampling (LHS) \cite{mckay2000comparison} which is the sampling strategy used in the first paper re-popularizing PINNs \cite{raissi2019physics}.

Random sampling methods such as LHS sample collocation points independently of the solution or domain. While these methods are shown to be effective, methods of sampling collocation points which take specific solution or domain features into consideration when sampling are known as adaptive point sampling strategies. This covers a wide spectrum of strategies. Adaptive sampling strategies may include sampling strategies which add collocation points during training \cite{lu2021deepxde}, resample the entire collocation point distribution during training \cite{wu2023comprehensive}, or moving certain collocation points throughout the training process \cite{hou2023enhancing}.

Frequently used adaptive sampling strategies in PINNs are residual-based adaptive strategies. These strategies use the PDE residual of collocation points as the factor which informs the sampling strategy. We give a brief overview of three of these strategies, where more details can be found in \cite{lu2021deepxde, wu2023comprehensive}.

Residual-based adaptive refinement (RAR) \cite{lu2021deepxde} is the first of these strategies. The goal of RAR is to improve the original collocation point distribution by adding points to areas of large PDE residual. After training for \(i\) epochs, a new sample of collocation points is generated. Then the PDE residual of the new points are calculated, with the \(n\) points with the largest residual added to the original distribution. Here, \(i, n\) are some predetermined number of iterations and collocation points, respectively.

Residual-based adaptive distribution (RAD) \cite{wu2023comprehensive} is a strategy in which the entire collocation point distribution is resampled every \(i\) epochs. This is done by first sampling a larger collocation point sample than the original distribution randomly. Then from this distribution we sample points to the same size of the original distribution by the probability density function
\begin{equation}\label{eq:radpdf}
    p(x) \propto \frac{\epsilon(x)}{\mathbb{E}[\epsilon(x)]} + c
\end{equation}
which is computed for every collocation point \(x\). Here, \( \epsilon(x)\) refers to the PDE residual, where \(\epsilon(x) = |\Delta(x; u_{(n)})|\). Hyperparameters \(k, c\) are chosen based the desired weighting of the probability distribution function to the collocation points with higher PDE residuals.

Residual-based adaptive refinement with distribution (RAR-D) \cite{wu2023comprehensive} is based on combining these two approaches. In this, after every \(i\) epochs of training, we add \(n\) new collocation points to the original distribution, similar to RAR. However, we add points not based on the PDE residual, but determined by the probability density function described by Eqn.~\eqref{eq:radpdf}.

\section{Physics-informed neural networks for differential equations (PinnDE)}\label{3}

Physics-informed neural networks for differential equations (PinnDE) is an open-source library in Python providing the ability to solve differential equations with both physics-informed neural networks (PINNs) and deep operator networks (DeepONets). PinnDE is built using TensorFlow \cite{tensorflow2015-whitepaper} and JAX \cite{jax2018github}, serving as the backend, two popular software packages for deep learning. Functionality to provide multiple backend and other deep learning frameworks such as PyTorch for use with PinnDE is in development and will be added in a future version.

The goal of PinnDE is to provide a user-friendly interface for solving systems of differential equations which can be easily shared with collaborators even if they do not have experience working with machine learning in Python. Many alternative packages for solving differential equations with PINNs or DeepONets, such as DeepXDE \cite{lu2021deepxde} and PINA \cite{coscia2023physics}, are powerful packages in this field. They provide a high level of customization and control in how a specific problem is being solved. This naturally leads to a higher amount of software needing to be written and understood by a user, which when sharing with collaborators can lead to difficulties if not everyone understands the package to the same degree. PinnDE is a tool in which we aim to bridge this gap between functionality and ease-of-use, stripping away complexity from the code resulting in an easy to read interface non-specialists can understand. This package has users create small, modular objects which represent components of a differential equation problem, which then requires just minimal interfacing with the package albeit still being able to have access to a wide range of applications, without the need to do any low-level manual implementations.

In these next sections we outline the overall structure of PinnDE, providing the current extent of this packages' capabilities. We then show how the process of solving a system of differential equations can be done using PinnDE.

\subsection{Overview}

\begin{figure}[!ht]
    \centering
    \includegraphics[width=0.8\linewidth]{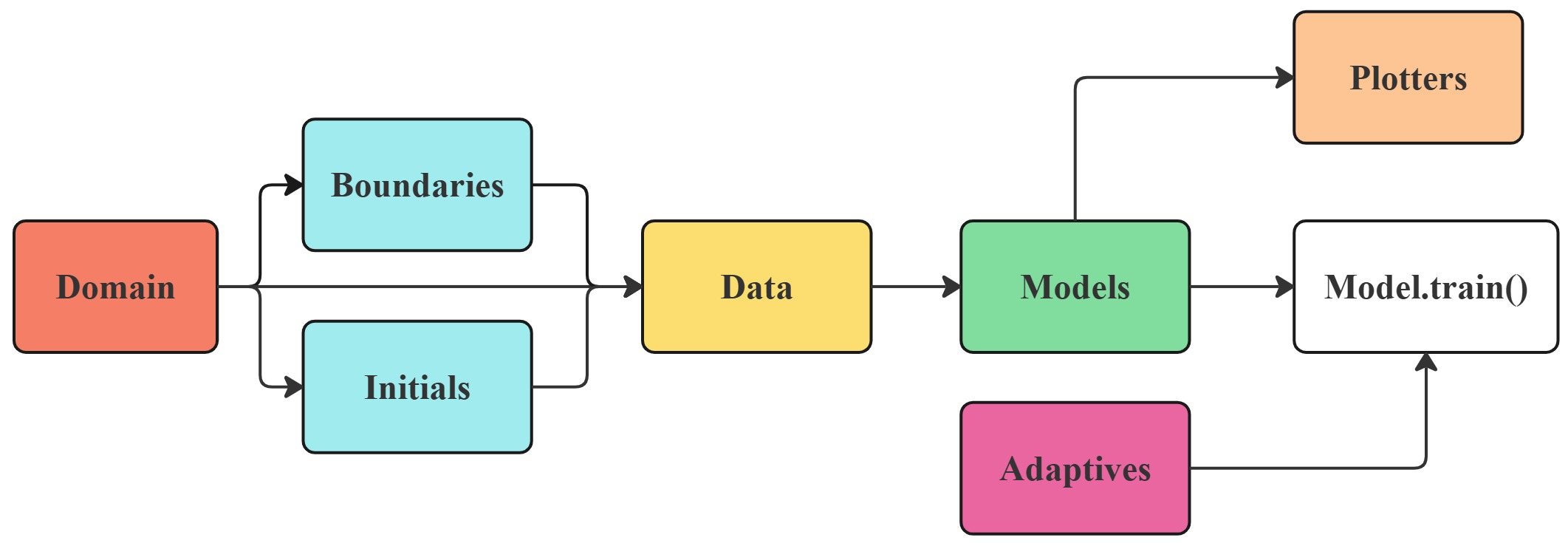}
    \caption{Flow chart of the module structure of PinnDE}
    \label{PinnDEchart}
\end{figure}

Figure \ref{PinnDEchart} shows the data flow of main modules involved in solving a system of differential equations using PinnDE. Each individual module gives the user a small amount of requirements to specify, all of which are related to the problem specification itself as well as to the basic neural network architecture and optimization procedure, but as will be demonstrated below still does not take away from the users' ability to interact with all further parts of the solution process. 

A user first initializes the domain of their problem using a single function from the \textbf{Domain} module. Next, the boundary conditions of the problem are initialized with the \textbf{Boundaries} module. If the problem is time-dependent and involves initial conditions, the initial conditions are defined using a single function from the \textbf{Initials} module. Then a single function in the \textbf{Data} module is being called corresponding to the system of differential equations one wishes to solve. This function call creates a data object, which is what is passed to a function call in the \textbf{Model} module to create the desired model to solve the specified differential equations. As a next step one calls the \verb|train()| on one's model, invoking the training routine for the specific problem with desired functionality (hard constraining, adaptive collocation points, meta-learned adaptivity, etc). Lastly, PinnDE then provides a plotting module to quickly plot losses and visualization of results on specific dimension equations.

In these next sections we give a brief overview of the domains, boundaries, data, which models and what adaptive point sampling strategies are currently available in PinnDE, specifying limitations in implementation if applicable. We only outline the general ability of PinnDE, a full API and tutorials can be found at \url{https://pinnde.readthedocs.io/en/latest/}.

\subsubsection{Domains} \label{3.1.1}
The domain module is the first which must be initialized when using PinnDE. Currently we provide two general types of domains implemented, \(d\) dimensional hyper-rectangles and \(d\) dimensional ellipsoids. Both of these domains allow specifying a domain in \(d\) dimensions, denoted in PinnDE as \(x1, \dots, xd\). Both of these have a spatial domain function, as well as a temporal domain counterpart, which is used when defining a time dependent system of partial differential equations, which  allows the time dimension, denoted \(t\) in PinnDE, to be used. This enables solving \((1+d)\)-dimensional evolution equations or \(d\)-dimensional boundary value problems. We also provide a general API for boundary functions with the abstract \verb|boundaries| class, which when implemented allow users to define custom-domains which will work with the PinnDE architecture.

\subsubsection{Initial and Boundaries} \label{3.1.2}

The \textbf{Boundaries} module provides boundary generation functions for ordinary and partial differential equations. The boundaries currently available to be used for partial differential equations are \textit{Periodic}, \textit{Dirichlet}, and \textit{Neumann}. We further provide the boundary \textit{ODEicbc}, which is designed to be used to specify initial or boundary conditions for systems of ordinary differential equations. All boundary functions require a domain to be passed in for which the boundary is being defined. These boundaries can be specified for any \((1+d)\)-dimensional system of evolution equations or \(d\)-dimensional boundary value problems. Specific implementation of the boundary values depends on the domain being described. 

PinnDE also offers both soft and hard constraining for a selection of these boundaries in specific dimensions. The availability of what PinnDE provides is further described in Section \ref{3.1.4} where the \textbf{Models} of PinnDE are described. Here we give some examples of specific implementations of how hard constraining of a boundary (temporal or spatial) is done in PinnDE.

Periodic boundary conditions over \(n\) spatial dimensions \((x_{1}, \dots,x_{n})\) of a differential equation over the hyper-rectangle \((x_{1}, \dots,x_{n}) \in [x_{\rm l_{1}}, x_{\rm r_{1}}] \times \dots \times [x_{\rm l_{n}}, x_{\rm r_{n}}]\) are enforced using a coordinate transform layer in the neural network itself, by using the transform \(f(x_{i})=(\cos 2\pi x_{i}/(x_{\rm r_{i}}-x_{\rm l_{i}}), \sin 2\pi x_{i}/(x_{\rm r_{i}}-x_{\rm l_{i}}))\) applied to each spatial input \(x_{i}\). See~\cite{bihlo2022physics} for further details.

For constraining a periodic boundary condition of a \((1+1)\)-dimensional PDE in variables \((t, x)\) over a spatio-temporal domain \([t_{\rm 0}, t_{\rm f}] \times \Omega\), \(\Omega =[x_{\rm l}, x_{\rm r}], x_{\rm l}, x_{\rm r} \in \mathbb{R}\), we only have to hard-constrain the initial conditions as periodic boundaries can be enforced exactly using a coordinate transform layer described above. For hard constraining the initial condition itself, we assume that the equation is of second order in \(t\). Then, we can include the initial condition as a hard constraint in the neural network surrogate solution \(u^\theta(t,x)\) using the ansatz
\begin{equation*}
    u^\theta(t,x) = u_0(x) + \frac{\partial u}{\partial t}\bigg|_{t=t_{\rm 0}}(x)\left(t-t_{\rm 0}\right) + \mathcal{N}^\theta(t,x)\left(\frac{t-t_{\rm 0}}{t_{\rm f} - t_{\rm 0}}\right)^2,
\end{equation*}
Where \(u_0\) and \(\frac{\partial u}{\partial t}\big|_{t=t_{\rm 0}}\) represent the initial conditions at \(t=t_{\rm 0}\), and \(\mathcal{N}^\theta(t,x)\) is the neural network output. Multiple variations of possible hard constraining formulas for this problem can be created which enforce the desired initial conditions. We found this particular hard constraint ansatz to be effective for many problems we have considered, and it is therefore used in PinnDE as a default hard constraint. However more hard constraints are available in PinnDE, including custom hard constraints.

We also present the hard constraint for a two-dimensional Dirichlet boundary-value problem for a PDE in variables \((x,y)\) over the rectangular domain \(\Omega = [x_{\rm l},x_{\rm r}] \times [y_{\rm l},y_{\rm u}]\), with \(x_{\rm l}, x_{\rm r}, y_{\rm l}, y_{\rm u} \in \mathbb{R}\). The approximate solution \(u^\theta(x,y)\) is given by the equation
\begin{equation*}
    u^\theta(t,x) = A(x,y) + x^*(1-x^*)y^*(1-y^*) \mathcal{N}^\theta(x,y),
\end{equation*}
where we have
\begin{align*}
    & A(x,y) = (1-x^*)f_{x_{\rm l}}(y) + x^*f_{x_{\rm r}}(y)+(1-y^*)[g_{y_{\rm l}}(x)-[(1-x^*)g_{y_{\rm l}}(x_{\rm l})+x^*g_{y_{\rm l}}(x_{\rm r})]]&\\
    &\quad \quad \quad \quad + y^*[g_{y_{\rm r}}(x) - [(1-x^*)g_{y_{\rm u}}(x_{\rm l}) + x^*g_{y_{\rm u}}(x_{\rm r})]],&
\end{align*}
where \(f_{x_{\rm l}}, f_{x_{\rm r}}, g_{y_{\rm l}}, g_{y_{\rm u}}\) represent some boundary functions \(f(x_{\rm l}, y), f(x_{\rm r}, y), g(x, y_{\rm l}), g(x, y_{\rm u})\) respectively, and 
\begin{equation*}
    x^* = \frac{x-x_{\rm l}}{x_{\rm r} - x_{\rm l}}, \quad  y^* = \frac{y-y_{\rm l}}{y_{\rm u} - y_{\rm l}}.
\end{equation*}
This formula enforces the specific boundary conditions for all output values of the neural network. This is based on what is described in \cite{lagaris1998artificial} and is the default hard constraint for two-dimensional spatial problems over a rectangular domain in PinnDE. Other alternative hard constraints do exist in PinnDE if applicable to the scenario, and the ability for user's to create custom hard constraints also exist.

Hard constraints can be designed for general domains, however the particular domain features impact the hard constraining function, making them cumbersome to design. In PinnDE, we give designed hard constraints pre-made for the most popular use cases of domain-boundary combinations. We currently provide built-in hard constraints for rectangular domains. We provide hard constraints in the following cases: Single ODEs, single \((1+1)\)-dimensional PDEs with periodic boundaries, and single \(2\)-dimensional PDEs with periodic and Dirichlet boundaries. However, PinnDE is built for custom hard constraints to be able to be created and added to the network, allowing for any user-specific hard constraint to be implemented on the network. In doing this, we cover functionality for most use cases while still having the possibility for customization for any problem.

\subsubsection{Data} \label{3.1.3}

The \textbf{Data} module of PinnDE combines and creates sampled collocation points for training. Each model has its own specific data object, with a time data counterpart to incorporate a time dimension when training for spatio-temporal equations. The data module always takes the \textbf{Domain} and \textbf{Boundaries} as arguments. If using a time domain, it also must take in an \textbf{Initials}. Then model specific information may be passed depending on the specific data being created. For example, DeepONet data objects need the number of sensor points to sample, and inverse PINN data objects need the sampled data over the problem domain and all functions being solved for. This is then the final step before defining the model.

\subsubsection{Models} \label{3.1.4}
PinnDE currently provides the ability to use PINNs and DeepONets for solving systems of differential equations. For PINNs, we provide standard forward PINNs as well as inverse PINNs for solving inverse problems. The number of layers and nodes can both be selected within each model function. The outer and inner activation function used within each model can be chosen by the user as well. The user also has the choice to either soft or hard constrain the network. We employ multi-layer perceptron for all neural network solution surrogates, and implement DeepONets using the architecture proposed in \cite{lu2019deeponet}. Both physics-informed neural networks and physics-informed deep operator networks are trained by minimizing the composite loss function~\eqref{eqn_2} to approximate the solution (or solution operator) of a system of differential equations. Derivative approximations are computed using automatic differentiation~\cite{baydin2018automatic}. Longer time integrations using DeepONets via time-stepping is realized using the procedure of~\cite{wang2023long}, which iteratively applies the learned solution operator to its own output to advance the solution for arbitrary time intervals, similar as done in standard numerical methods.

\subsubsection{Adaptive} \label{3.1.5}

When calling \verb|train()| on a model from PinnDE, the user is allowed to pass in an adaptive point sampling strategy created from the \textbf{Adaptive}'s module. We currently provide three different adaptive sampling strategies which can be employed without needing to do any manual implementation: residual-based adaptive refinement, residual-based adaptive distribution, and residual-based adaptive refinement with distribution (RAR, RAD, RAR-D). These strategies are based on the research reported in \cite{lu2021deepxde, wu2023comprehensive}, and were shortly summarized in Section \ref{2.4}. These strategies can be used with both forward and inverse PINNs, and can also be applied in combination with L-BFGS optimization.

\subsection{Usage}

In this section we give the simple outline for how to use PinnDE. We describe what data must be declared when solving a system of differential equations. We give an example of the standard process solving a single partial differential equation. The process outline is described in Table \ref{tab:workflow}.
\begin{table}[H]
    \centering
    \caption{Workflow to solve PDE}
    \begin{tabular}{l l}
        \hline\hline
        Step 1 & Use the \textbf{Domain} module to define the domain for the independent variables \\
        Step 2 & Define the boundary functions as Python \verb|lambda| functions, \\
        & as well as number of sample points to use on boundary \\
        Step 3 & Use the \textbf{Boundaries} module to set up the desired boundary \\
        Step 4 & Define all initial conditions as Python \verb|lambda| functions, number of\\
        & initial value points, and use the \textbf{Initials} module to set up initial conditions \\
        Step 5 & Define the \textbf{Data} object corresponding to the model being used\\ 
         Step 6 & Optionally declare number of layers of network, nodes of network,\\
         & and constraint of network to interface with model\\
         Step 7& Define the equation to solve, and use \textbf{Model} module and input data to solve equation\\
         Step 8& Call \verb|train(epochs)| on returned model to train the network\\
         Step 9& Use trained model to plot data with given functions or take data from class\\
         & to use independently
    \end{tabular}
    \label{tab:workflow}
\end{table}

Example code that follows this structure is given in Appendix \ref{A}, where the code for the following examples in Section \ref{4} is presented. Many more tutorials are given along with the API in the linked documentation above.

\section{Examples} \label{4}

In this section we present a few short examples of how PinnDE can be used to solve systems of ordinary differential equations, and systems of partial differential equations using the different methods implemented in PinnDE. We provide a concise description of each problem and a comparison of each trained model against the analytical solution or a numerical reference solution obtained using a standard numerical integrator using the mean squared error as comparison metric. All code for these examples is provided in Appendix \ref{A}.

\subsection{Rigid body} \label{4.1}

We first consider the Euler equations for the free rigid body~\cite{holm2009geometric}, a system of ordinary differential equations, with a physics-informed network which uses soft constrained initial conditions. Specifically, we solve the system of order ordinary differential equations in angular momentum representation
\begin{subequations}\label{eq:ODESystem}
\begin{align}
    \frac{dw_{1}}{dt} &=\frac{I_{2} - I_{3}}{I_{2}I_{3}}w_{2}w_{3},\\
    \frac{dw_{2}}{dt} &=\frac{I_{3} - I_{1}}{I_{1}I_{3}}w_{1}w_{3},\\
    \frac{dw_{3}}{dt} &=\frac{I_{1} - I_{2}}{I_{1}I_{2}}w_{1}w_{2},
\end{align}
\end{subequations}
where \(w = (w_{1}, w_{2}, w_{3})\) is the angular momentum of the rigid body, and \(I_{1}, I_{2}, I_{3}\) are the principle moments of inertia. For this benchmark we use \(I_{1}=0.2, I_{2}=0.3, I_{3}=0.4\), and the initial condition \(w = (1, 1, 1)\). We solve this initial value problem over the interval \(t \in [0,2.5]\).

\begin{figure}[!ht]
     \centering
    \includegraphics[width=1\linewidth]{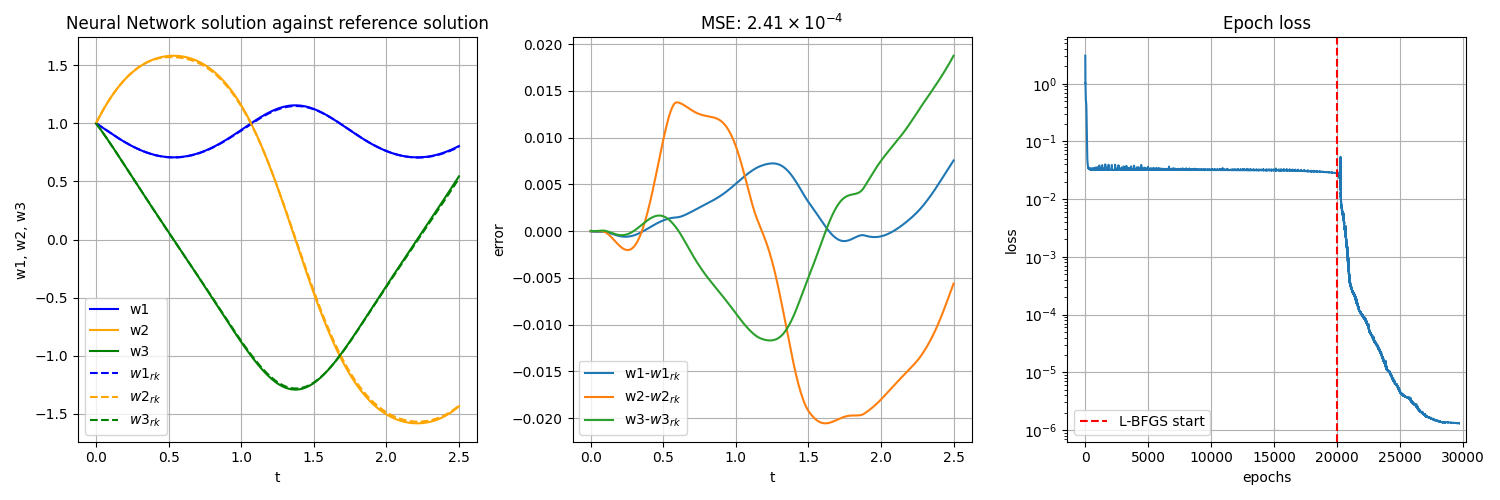}
    \caption{Solution predicted using a PINN against the numerical solution for the rigid body system~\eqref{eq:ODESystem} \textit{(left)}, the difference of neural network solution and numerical solution \textit{(middle)}, and the time series of loss from training the network \textit{(right)}.}
    \label{fig:ODE}
\end{figure}

Figure \ref{fig:ODE} presents the neural network solution against a numerical solution as well as the time series of the loss for training the network. The standard numerical reference solution for this problem was computed using the standard fourth-order Runge--Kutta method.
The specific architecture used was a PINN with 4 hidden layers with 60 nodes per layer, using the hyperbolic tangent as activation function, and Adam and L-BFGS for fine-tuning was chosen as the optimizers. We use 3000 collocation points across the domain for the network to learn the solution, sampled with Latin hypercube sampling. We train this network for 20000 epochs using Adam, then 5000 iterations using L-BFGS, implementing a two-step training routine. The start of L-BFGS optimization during training is highlighted in Fig.~\ref{fig:ODE}. We stop training when the epoch loss converges towards the solution. For the following examples, we will also train until a minimum loss has been reached.

The code for solving this system of equations can be found in Appendix \ref{A.1}.

\subsection{Lorenz model} \label{4.2}

We next use PinnDE to solve the Lorenz--1960 model, a system of ordinary differential equations, with a deep operator network which uses soft-constrained initial conditions. The dynamical system is
\begin{subequations}\label{eq:LorenzSystem}
\begin{align}
    \frac{dx}{dt} &= kl(\frac{1}{k^{2}+l^{2}} - \frac{1}{k^{2}})yz,\\
    \frac{dy}{dt} &= kl(\frac{1}{l^{2}} - \frac{1}{k^{2}+l^{2}})xz,\\
    \frac{dz}{dt} &=\frac{kl}{2}(\frac{1}{k^{2}} - \frac{1}{l^{2}})xy,
\end{align}
\end{subequations}
where in this benchmark we chose \(k=2, l=1\), and we choose initial conditions \(x(0)=0.5, y(0)=0.75, z(0)=1\). We solve this initial value problem over the interval \(t \in [0,1]\).

\begin{figure}[!ht]
     \centering
    \includegraphics[width=1\linewidth]{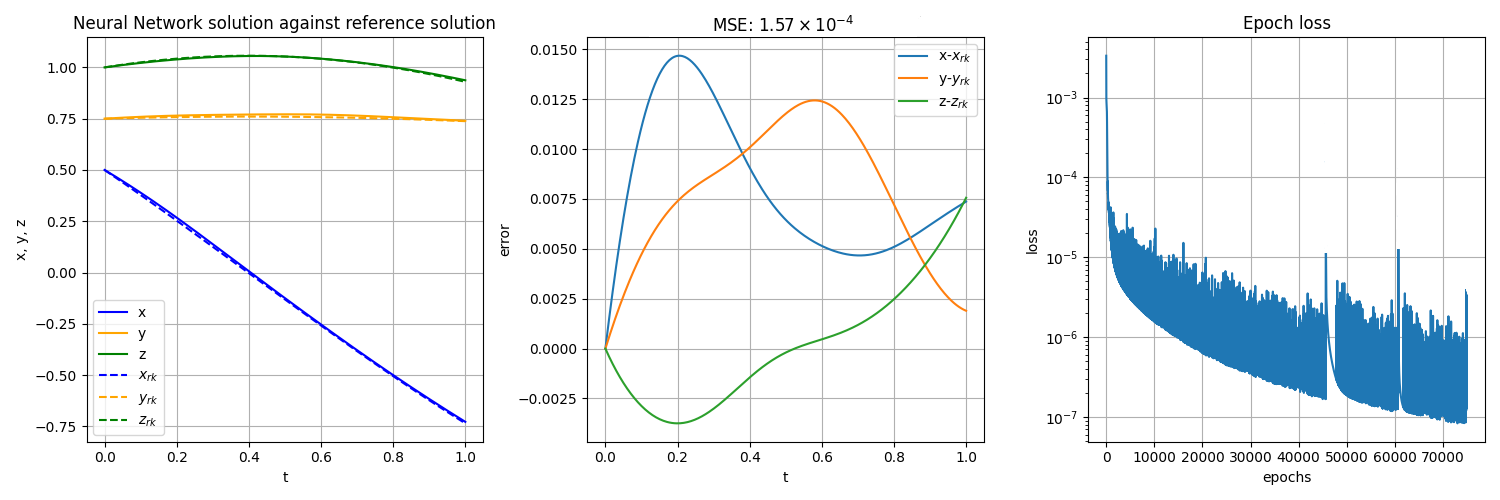}
    \caption{Solution predicted using a DeepONet against the numerical solution for the Lorenz system~\eqref{eq:LorenzSystem} \textit{(left)}, the difference of neural network solution and numerical solution \textit{(middle)}, and the time series of loss from training the network \textit{(right)}.}
    \label{fig:Lorenz}
\end{figure}

Figure \ref{fig:Lorenz} presents the neural network solution against a numerical solution as well as the time series of the loss for training the network. The standard numerical reference solution for this problem was computed using the Runge--Kutta method.
The specific architecture used was a DeepONet with both branch net and trunk net being multi-layer perceptrons having 4 hidden layers with 40 nodes per layer, using the hyperbolic tangent as activation function, and Adam as the chosen optimizer. We use 3000 collocation points across the domain for the network to learn the solution, sampled with Latin hypercube sampling. We sample 5000 different initial conditions for the DeepONet's branch network in the range of \(x(0), y(0), z(0)\in[-2, 2]\). We train this network for 75000 epochs using Adam. This is when the epoch loss converges towards its minimum.

\begin{figure}[!ht]
     \centering
    \includegraphics[width=1\linewidth]{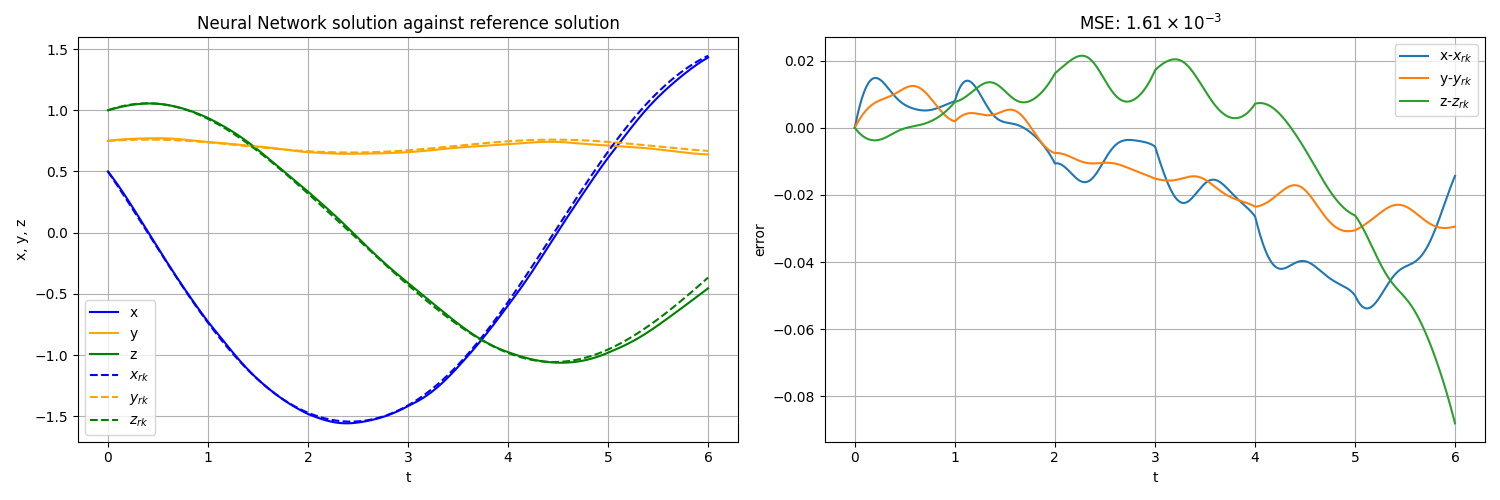}
    \caption{Time-stepped solution predicted by the trained DeepONet against the numerical reference solution obtained using a fourth-order Runge--Kutta method for system~\eqref{eq:LorenzSystem} \textit{(left)}, the difference of neural network solution and numerical solution \textit{(right)}.}
    \label{fig:Lorenztimestep}
\end{figure}

We further demonstrate the time-stepping ability using a DeepONet in PinnDE in Fig.~\ref{fig:Lorenztimestep}. We time-step the trained DeepONet for 6 steps, increasing the temporal domain to \([0, 6]\), and give the error of the neural network solution against the corresponding Runge--Kutta numerical solution. The learned neural network solution operator demonstrates the ability to time-step with a consistently low error.

The code for solving this system of equations can be found in Appendix \ref{A.2}.

\subsection{Dam break problem} \label{4.3}

\begin{figure}[!ht]
    \centering
    \begin{subfigure}{1\textwidth}
       \includegraphics[width=1\linewidth]{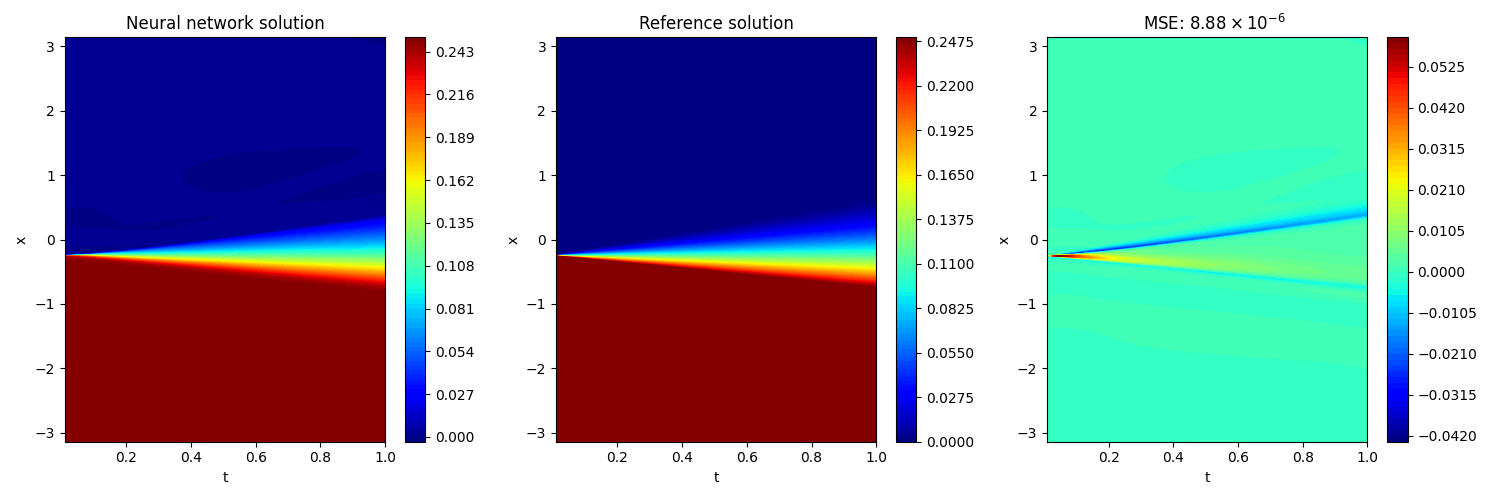} 
    \end{subfigure}
    \begin{subfigure}{1\textwidth}
       \includegraphics[width=1\linewidth]{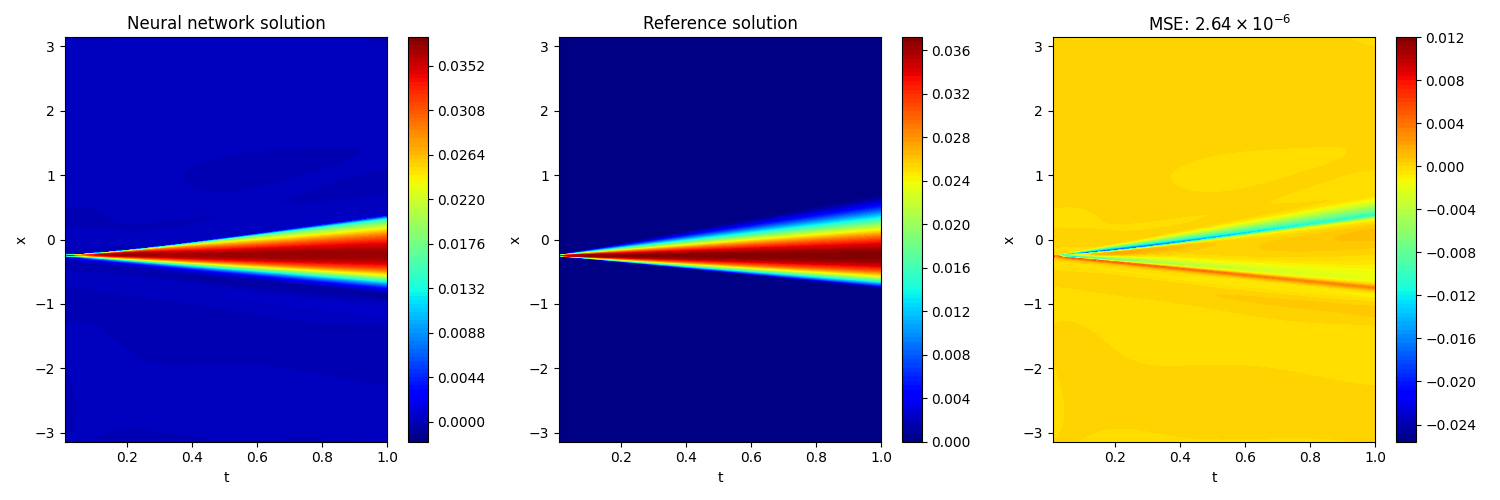} 
    \end{subfigure}
    \caption{Solution predictions using a PINN for the dam break problem~\eqref{eq:SWE} \textit{(left)}, the analytical solution \textit{(middle)}, and the difference between them \textit{(right)}. \textit{Top:} Water height \(h(t, x)\). \textit{Bottom:} Momentum \(h(t, x)u(t, x)\).}
    \label{fig:dam}
\end{figure}

We next use PinnDE to solve the classical one-dimensional dam break problem with flat bottom topography for the shallow-water equations. We use soft-constrained initial conditions and Dirichlet boundary conditions. Specifically, we solve the one-dimensional shallow water equations given by
\begin{subequations}\label{eq:SWE}
\begin{align}
    &\frac{\partial h}{\partial t} + \frac{\partial (hu)}{\partial x} = 0,\\
    &\frac{\partial (hu)}{\partial t} + \frac{\partial}{\partial x}\left(hu^{2}+\frac{1}{2}gh^{2}\right) = 0,
\end{align}
\end{subequations}
where \(h(t, x)\) represents the fluid depth and \(u(t, x)\) represents the fluid velocity. The analytical solution for this problem is given by
\begin{align*}
    &gh_{a}(t, x) = \begin{cases}
        a_{0}^{2} & \text{if } x-x_{0} < -a_{0}t \\
        \frac{1}{9}(2a_{0} - (x-x_{0})/t)^{2} & \text{if } -a_{0}t < x-x_{0} < 2a_{0}t \\
        0 &\text{if } x-x_{0} > 2a_{0}t
    \end{cases}\\
    &u_{a}(t, x) = \begin{cases}
        0 & \text{if } x-x_{0} < -a_{0}t \\
        \frac{2}{3}(a_{0} + (x-x_{0})/t)^{2} & \text{if } -a_{0}t < x-x_{0} < 2a_{0}t \\
        0 &\text{if } x-x_{0} > 2a_{0}t
    \end{cases}
\end{align*}
where \(a_{0} = \sqrt{gh_{0}}\), see~\cite{bokhove2005flooding} for further details. This analytical solution gives the initial and boundary conditions. In this benchmark, we use initial parameters \(h_{0} = 0.75, x_{0} = -0.25\), and we set \(g=1\). We consider this benchmark over the spatio-temporal domain \((t, x) \in [0.01, 1] \times [-\pi, \pi]\). We compare the neural network solution to the analytical solution for both \(h(t, x)\) and \(h(t,x)u(t,x)\) in Figure \ref{fig:dam}.

For this example we used a PINN with 4 hidden layers with 60 nodes per layer, the hyperbolic tangent activation function, and trained the neural network using the Adam optimizer. A total of 1000 initial value collocation points were used for the network to learn the initial condition, 2000 boundary collocation points, and 15000 collocation points were used across the spatio-temporal domain for the network to learn the solution, both sampled using Latin hypercube sampling. The minimum of the loss was reached after 35000 epochs of training.

The code for solving this equation can be found in Appendix \ref{A.3}.

\subsection{Burgers' equation} \label{4.4}

\begin{figure}[!ht]
    \centering
    \includegraphics[width=1\linewidth]{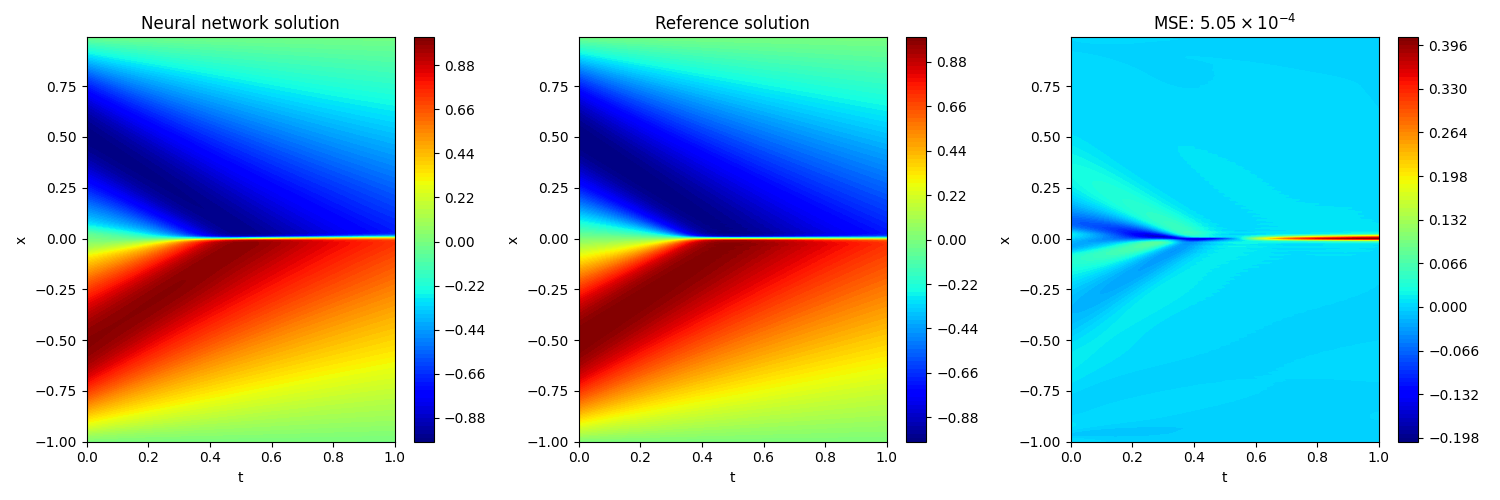}
    \caption{Solution predicted using a PINN for Burgers' equation~\eqref{eq:BurgersEquation} \textit{(left)}, the numerical reference solution \textit{(middle)}, and the mean squared error between them \textit{(right)}.}
    \label{fig:Burgers}
\end{figure}

We next use PinnDE to solve Burgers' equation with a PINN which uses soft-constrained initial and boundary conditions. We solve this equation using periodic boundary values. We also use a residual-based adaptive distribution (RAD) adaptive sampling strategy. In particular, we consider
\begin{equation}\label{eq:BurgersEquation}
    \frac{\partial u}{\partial t} + u\frac{\partial u}{\partial x} - \nu\frac{\partial^{2} u}{\partial x^{2}}=0,\qquad \nu=\frac{0.01}{\pi}.
\end{equation}
 We solve this equation over the square domain \( (t,x) \in [0,1] \times [-1, 1]\). We compare the neural network solution to a numerical reference solution in Figure \ref{fig:Burgers}. The numerical reference solution uses a spectral discretization in space combined with the method of lines based on the Runge--Kutta integrator for time-stepping.
For this example we used a PINN with 4 hidden layers with 60 nodes per layer, the hyperbolic tangent activation function, and trained the neural network using a two step training routine. The Adam optimizer was used to train the network for 5000 epochs. A total of 400 initial value collocation points were used for the network to learn the initial condition, and 2000 collocation points were used across the spatial domain for the network to learn the solution, all sampled using Latin hypercube sampling. 

\begin{figure}[H]
     \centering
     \begin{subfigure}{0.49\textwidth}
         \centering
         \includegraphics[width=\textwidth]{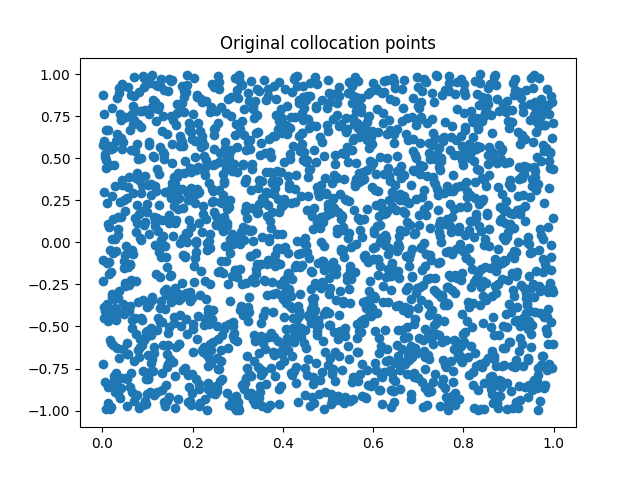}
     \end{subfigure}
     \hfill
     \begin{subfigure}{0.49\textwidth}
         \centering
         \includegraphics[width=\textwidth]{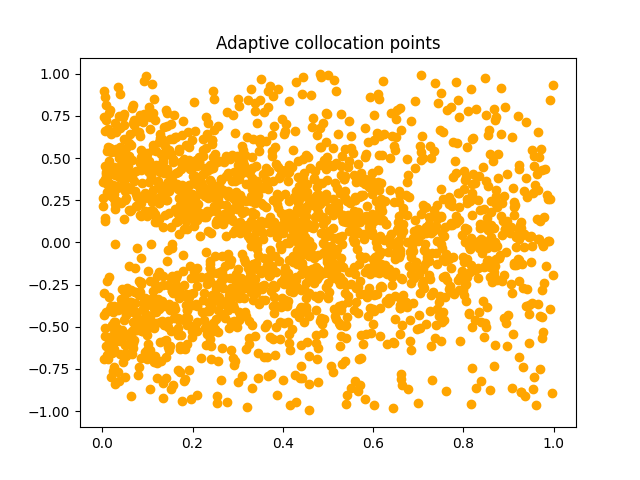}
     \end{subfigure}
        \caption{Original collocation points sampled using Latin hypercube sampling \textit{(right)} and the resampled distribution by RAD \textit{(left)}.}
        \label{fig:Burgerspoints}
\end{figure}

We used residual-based adaptive distribution (RAD) to train this equation. This strategy resamples the distribution of collocation points during training to align with areas of larger PDE residual. For Burgers' equation, this is particularly effective as we have a steep developing shock which to resolve requires locally more collocation points. We resample the distribution after 3000 epochs of training. We sample 10000 collocation points across the domain, and sample 2000 new collocation points from the probability distribution function described in Section \ref{2.4}, with hyperparameters \(k=3, c=0\). We compare these points in Fig.~\ref{fig:Burgerspoints}.

The code for solving this equation can be found in Appendix \ref{A.4}.

\subsection{Heat Equation} \label{4.5}

\begin{figure}[!ht]
\centering
\begin{subfigure}{0.32\textwidth}
   \includegraphics[width=1\linewidth]{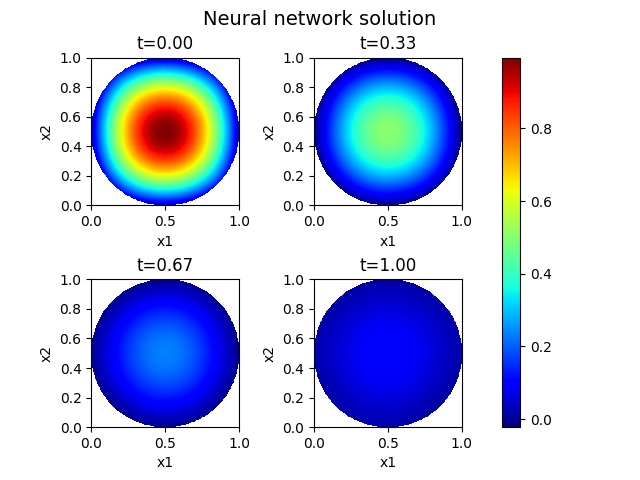} 
\end{subfigure}
\begin{subfigure}{0.32\textwidth}
   \includegraphics[width=1\linewidth]{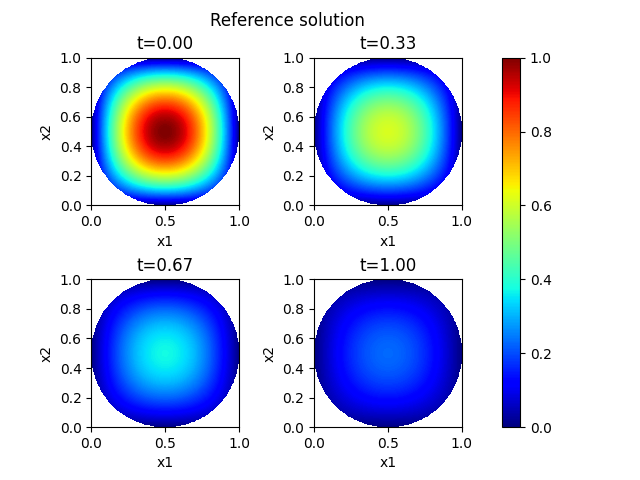} 
\end{subfigure}
\begin{subfigure}{0.32\textwidth}
   \includegraphics[width=1\linewidth]{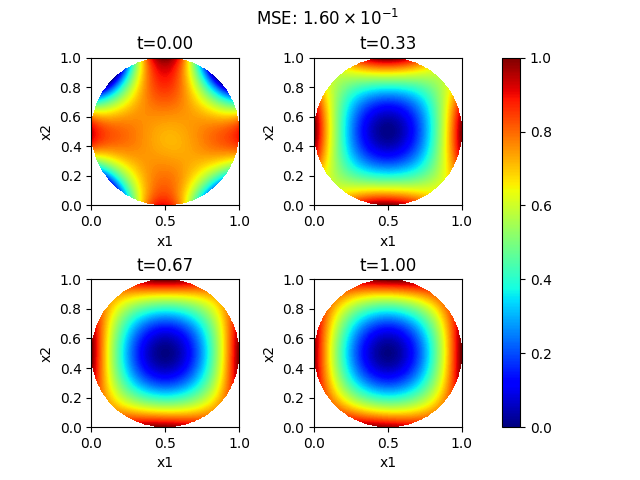} 
\end{subfigure}
\caption{Solution predicted by DeepONet for the heat equation~\eqref{eq:HeatEquation} \textit{(left)}, the analytical solution \textit{(middle)}, and the mean squared error between them \textit{(right)}.}
\label{fig:Heat}
\end{figure}

For our next example we use PinnDE to solve the \((1+2)\)-dimensional linear heat equation with a DeepONet which uses soft-constrained initial and boundary conditions, meaning we use a composite loss function consisting of the PDE, initial condition, and boundary condition losses. We solve this equation with Dirichlet boundary values. Specifically, we consider the equation
\begin{subequations}\label{eq:HeatEquation}
\begin{equation}
    \frac{\partial u}{\partial t} = \nu\left( \frac{\partial^2 u}{\partial x^2} + \frac{\partial^2 u}{\partial y^2}\right), \quad \nu=0.15,
\end{equation}
with initial condition
\begin{equation}
    u_0(x, y) = \sin\pi x \sin\pi y,
\end{equation}
and with Dirichlet boundary conditions
\begin{equation}
    u(t, x, y)=0 \quad\forall (t, x, y) \in \{ [0,1]\times\mathbb{R}^{2} | x^2 + y^2 = 1\}
\end{equation}
\end{subequations}
We solve this equation over the time interval \(t\in[0,1]\), on the spatial domain of the unit circle \((x, y)\in \{\mathbb{R}^{2} | x^2 + y^2 < 1\}\). We compare the neural network solution to the analytical solution \(u(t, x, y) = \exp(\nu\pi^2t) \sin\pi x \sin\pi y\) in Figure \ref{fig:Heat}.

For this example we trained a DeepONet with both branch net and trunk net being multi-layer perceptrons having 4 hidden layers with 60 nodes per layer, using the hyperbolic tangent activation function. We use Adam as our chosen optimizer. 600 initial value points were used to be sampled along \(t=0\) for the network to learn the initial condition. The soft-constrained boundary was sampled using 600 boundary value points for the neural network to learn the boundary conditions, and 12000 collocation points were used across the domain for the network to learn the solution itself, both sampled with Latin hypercube sampling. 
We sample a total of 1000 initial conditions from a class of truncated Fourier series with random coefficients to learn the solution operator. After training on this class of initial conditions, the operator network can be evaluated in a simple inference step for new initial conditions from the same initial condition class, an example of which is shown in Figure~\ref{fig:Heat}. We note here that different classes of initial conditions, such as Gaussian random fields~\cite{lu2019deeponet} could be considered as well for operator learning, and will be implemented in PinnDE in the future. We train this network for 4000 epochs which is when the minimum loss is reached

The code for solving this equation can be found in Appendix \ref{A.5}.

\subsection{Inverse linear advection speed} \label{4.6}

\begin{figure}[!ht]
    \centering
    \begin{subfigure}{1\textwidth}
       \includegraphics[width=1\linewidth]{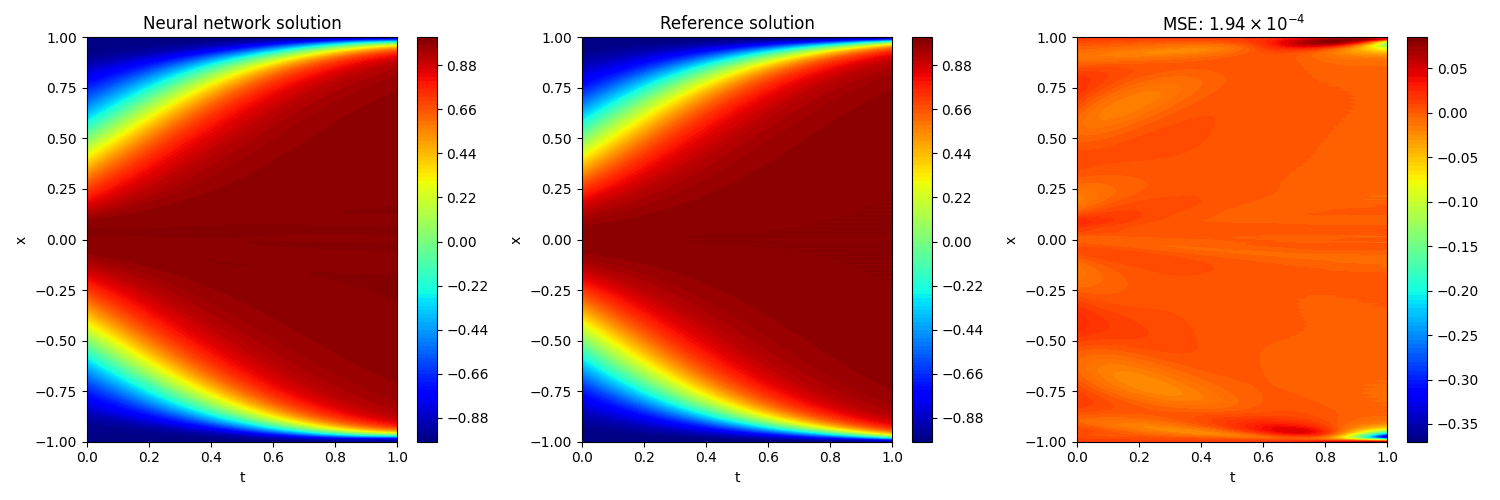} 
    \end{subfigure}
    \begin{subfigure}{1\textwidth}
       \includegraphics[width=1\linewidth]{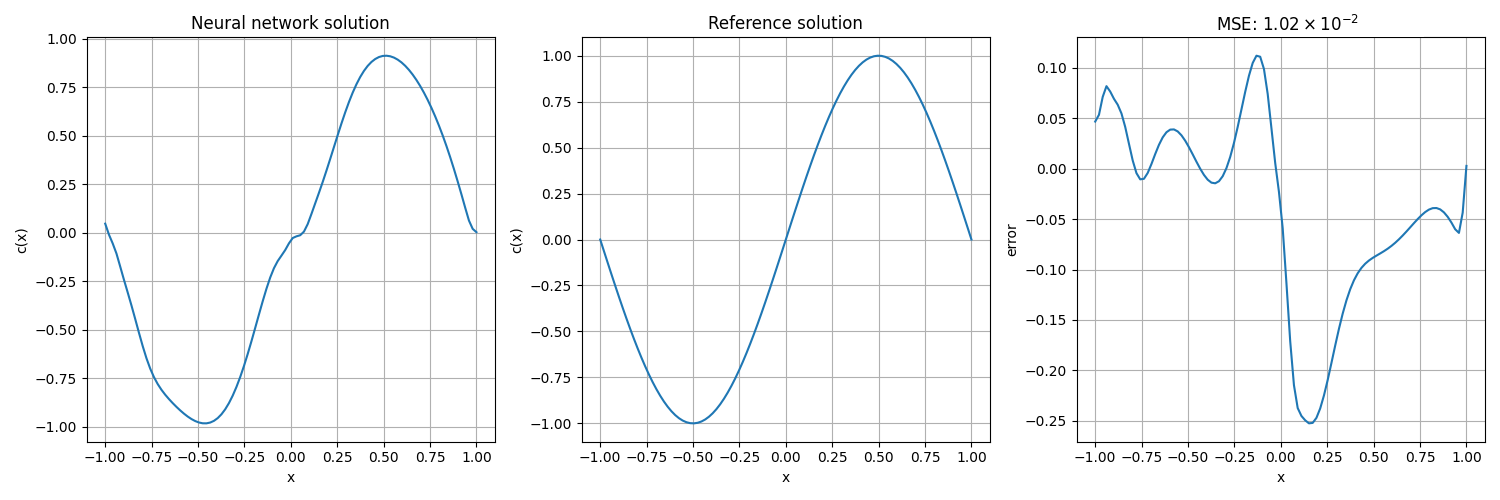} 
    \end{subfigure}
    \caption{Solution predictions using a PINN for the inverse linear advection speed problem~\eqref{eq:LinearAdvectionInv} \textit{(left)}, the numerical solution \textit{(middle)}, and the difference between them \textit{(right)}. \textit{Top:} \(u(t, x)\). \textit{Bottom:} Advection velocity \(c=\sin(\pi x)\).}
    \label{fig:LinAdvInv}
\end{figure}

We lastly use PinnDE to solve for an unknown advection speed in the linear advection equation with an inverse PINN which uses a soft-constrained initial condition. We use periodic boundary conditions, which in PinnDE are implemented at the neural network level as a hard constraint to imposes periodic boundaries on the independent spatial variable \(x\). We also use a residual-based adaptive refinement with distribution (RAR-D) adaptive sampling strategy. The linear advection with variable advection speed for our reference solution data is \(c=\sin (\pi x)\), and the equation is described as 
\begin{subequations}\label{eq:LinearAdvectionInv}
\begin{equation}
    \frac{\partial u}{\partial t} + c\frac{\partial u}{\partial x} = 0,
\end{equation}
where we use the initial condition
\begin{equation}
    u_0(x) = \cos\pi x.
\end{equation}
\end{subequations}
We solve this equation over the interval \(t\in[0,1]\), on the spatial domain \(x\in[-1,1]\). We compare the neural network solution to a numerical calculation of the solution in Figure \ref{fig:LinAdvInv}, showing the success of the trained model, as it gives small point-wise and overall mean-squared errors, respectively. We use a numerical solution here as no analytical solution for this equation exists.

\begin{figure}[!ht]
    \centering
    \includegraphics[width=1\linewidth]{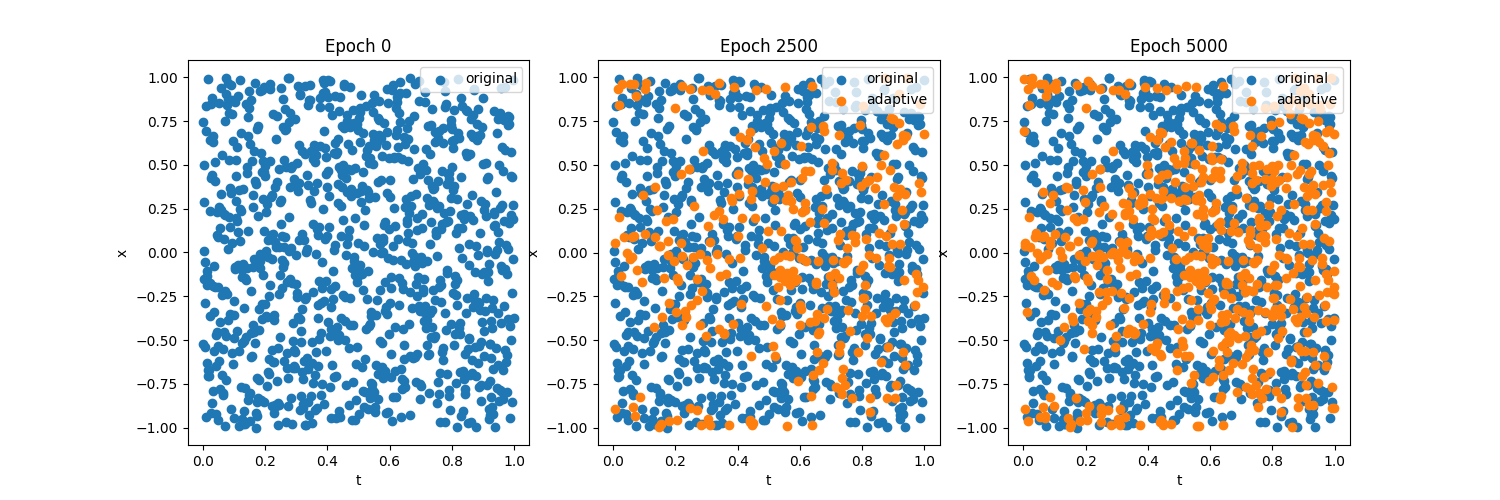}
    \caption{Original collocation points before training \textit{(left)}, the collocation points halfway through training at epoch 2500 \textit{(middle)}, and the final set of collocation points after 5000 epochs \textit{(right)}.}
    \label{fig:InvAdvecPoints}
\end{figure}

For this example we used a PINN with 4 hidden layers with 60 nodes per layer, the hyperbolic tangent activation function, and trained the neural network using the Adam optimizer. A total of 200 initial value collocation points were used for the network to learn the initial condition, and 1000 collocation points were used across the spatio-temporal domain for the network to learn the solution, both sampled using Latin hypercube sampling. We also sampled 500 points across the spatio-temporal domain over which we evaluate the analytical solution \(u\) as data points. These data points are then used to learn the advection speed \(c\). Our RAR-D sampling strategy added 30 collocation points every 250 epochs, with hyper-parameters \(k=3,c=0\) (see \cite{wu2023comprehensive} for more details). The evolution of collocation points throughout training is shown in Fig.~\ref{fig:InvAdvecPoints}. The minimum of the loss was reached after 5000 epochs of training.

The code for solving this equation can be found in Appendix \ref{A.6}.

\section{Conclusion} \label{5}

Deep learning-based approaches for solving differential equations have become an increasingly popular alternative for solving differential equations compared to methods based around numerical differentiation such as finite element methods, finite volume methods, and finite difference methods. PINNs and DeepONets are two methodologies of this deep learning-based approach. With these, we have a naturally meshless alternative to any of the classical numerical methods mentioned above. Deep learning-based approaches have the ability to use automatic differentiation to compute derivatives, without the need for any numerical discretizations to approximate derivatives. This streamlines the application of these methods across arbitrary domains, and by being able to incorporate physical constraints into the solving procedure, allow for them to solve problems in which classic numerical methods may struggle.

In this paper we presented PinnDE, an open-source software package in Python for PINN and DeepONet implementations for solving ordinary and partial differential equations. We reviewed the methodologies behind physics-informed neural networks for both forward and inverse problems, the DeepONet approach to operator learning, and adaptive point sampling strategies. We then introduced PinnDE and its functionality, and finally presented several worked examples to show the effectiveness of this package. 

Software packages giving the ability to use physics-informed machine learning approaches are becoming increasingly popular. In contrast to other existing packages, PinnDE stresses simplicity and requires minimal coding from the side of the user, thus alleviating the need of writing significant amounts of software to support the higher level of customization that other packages offer, while still providing many of the abilities that alternative packages offer. As such, PinnDE provides a simple interface where users with little experience can still write and understand the implementation of software written in this package. This gives researchers the ability to use this package with collaborators outside of the field who may be interested in this area of research but are lacking the skills to use more advanced packages. PinnDE also provides a tool for educators to use as the simple implementations give a low barrier of entry to new users. We also hope that by providing a package with such quick implementations of problems will promote further research and development of the field of scientific machine learning.

PinnDE is continuously being developed to integrate new advancements in research surrounding PINNs and DeepONets, while aiming to maintain a straightforward user experience. Some possible additions which will be implemented in future versions are more adaptive collocation point methods, other variants on DeepONet and PINN architectures, meta-learned optimization, the availability of different backend, and general geometries which differential equations can be solved over.

All code and link to documentation with guides can be found at: \url{https://github.com/JB55Matthews/PinnDE}

\begin{ack}
This research was undertaken thanks to funding from the Canada Research Chairs program and the NSERC Discovery Grant program.
\end{ack}

\section*{Author Contributions}
\textbf{Jason Matthews:} Writing - original draft,  Writing - review \& editing, software, conceptualization, methodology, visualization. \textbf{Alex Bihlo:} Writing - review \& editing, conceptualization, methodology, resources, supervision, funding acquisition.

\bibliographystyle{plain}
\bibliography{biblio}

\appendix
\section{Appendix} \label{A}
Here we provide the code corresponding to the examples presented in Sections [\ref{4.1}, \ref{4.2}, \ref{4.3}, \ref{4.4}, \ref{4.5}, \ref{4.6}].

\subsection{Rigid body code} \label{A.1}

The following is the code being used in Section \ref{4.1}.

\begin{verbatim}
    import src.pinnde as p
    import numpy as np

    re = p.domain.NRect(1, [0], [2.5])
    cond = p.boundaries.odeicbc(tre, [[1], [1], [1]], "ic")

    dat = p.data.pinndata(tre, cond, 3000, 300)

    eqns = ["(-5/6)*u2*u3 - u1x1", "(5/2)*u1*u3 - u2x1", 
                                    "(-5/3)*u1*u2 -u3x1"]
                                
    mymodel = p.models.pinn(dat, eqns)

    mymodel.train(20000)
    mymodel.train(5000, opt="lbfgs")

\end{verbatim}

\subsection{Lorenz model code} \label{A.2}

The numerical results presented in Section \ref{4.2} are obtained from the following code.

\begin{verbatim}
    import pinnde.legacy.ode_Solvers as ode_Solvers

    eqn1 = "(-1.6)*x*y - ut"
    eqn2 = "0.1*u*y - xt"
    eqn3 = "0.75*u*x - yt"
    eqns = [eqn1, eqn2, eqn3]
    orders = [1, 1, 1]
    inits = [[0.5], [0.75], [1]]
    t_bdry = [0,1]
    N_pde = 500
    sensor_range = [-2, 2]
    num_sensors = 3000
    epochs = 75000
    
    mymodel = ode_Solvers.solveODE_DeepONetSystem_IVP(eqns, orders,
            inits, t_bdry, N_pde, sensor_range, num_sensors, epochs)
            
    mymodel.timeStep(6)
    mymodel.plot_epoch_loss()
    mymodel.plot_solution_prediction()
\end{verbatim}

\subsection{Dam break problem code} \label{A.3}
To obtain the results presented in Section \ref{4.3}, the following code can be used.

\begin{verbatim}
    import pinnde as p
    import numpy as np
    import tensorflow as tf

    h_analytical = lambda t, x: tf.where(x+0.25<-0.5*t, 0.5**2,  
     tf.where(x+0.25<1*t, (1-(x+0.25)/t)**2/(9), 0.0))
    
    u_analytical = lambda t, x: tf.where(x+0.25<-0.5*t, 0.0, 
        tf.where(x+0.25<1*t, 2/3*(0.5+(x+0.25)/t), 0.0))

    h_init = lambda x: tf.where(x+0.25<-0.5*0.75, 0.5**2,
        tf.where(x+0.25<1*0.75, (1-(x+0.25)/0.75)**2/(9), 0.0))
    
    u_init = lambda x: tf.where(x+0.25<-0.5*0.75, 0.0, 
        tf.where(x+0.25<1*0.75, 2/3*(0.5+(x+0.25)/0.75), 0.0))

    tre = p.domain.Time_NRect(1, [-np.pi], [np.pi], [0.01, 1])
    
    bound = p.boundaries.dirichlet(tre, [h_analytical])
    
    inits = p.initials.initials(tre, [[h_init], [u_init]])
    
    dat = p.data.timepinndata(tre, bound, inits, 15000, 2000, 1000)

    eqn1 = "u1*u2t + u1t*u2 + 2*u1*u2*u2x1 + u1x1*u2**2 + u1*u1x1"
    eqn2 = "u1t + u1*u2x1 + u1x1*u2"
    mymodel = p.models.pinn(dat, [eqn1, eqn2])
    mymodel.train(35000)
\end{verbatim} 

\subsection{Burgers' equation code}  \label{A.4}

Here is the code which provides the results presented in Section \ref{4.4}.

\begin{verbatim}
    import pinnde as p
    import numpy as np
    import tensorflow as tf

    tre = p.domain.Time_NRect(1, [-1], [1], [0,1])

    bound = p.boundaries.periodic(tre)
    inits = p.initials.initials(tre, [lambda x: -np.sin(np.pi*x)])
    
    dat = p.data.timepinndata(tre, bound, inits, 2000, 600, 400)

    mymodel = p.models.pinn(dat, ["ut+u*ux1-(0.01/np.pi)*ux1x1"])
    rad = p.adaptives.RAD(3000, 3, 0, 5)
    mymodel.train(5000)    
\end{verbatim}

\subsection{Heat equation code}  \label{A.5}

The numerical results presented in Section \ref{4.5} are obtained from the following code.

\begin{verbatim}
    import pinnde as p
    import numpy as np
    import tensorflow as tf
    
    center = [0.5, 0.5]
    semilengths = [0.5, 0.5]
    timerange = [0, 1]
    dom = p.domain.Time_NEllipsoid(2, center, semilengths, timerange)
    
    bdryfunc = lambda t, x1, x2: 0+t*0
    bound = p.boundaries.dirichlet(tre, [bdryfunc])
    
    u0func = lambda x1, x2: tf.sin(np.pi*x1)*tf.sin(np.pi*x2)
    inits = p.initials.initials(dom, [u0func])
    
    dat = p.data.timedondata(dom, bound, inits, 12000, 600, 600, 1000)
    
    eqn = "0.15*ux1x1 + 0.15*ux2x2 - ut"
    mymodel = p.models.deeponet(dat, [eqn])
    mymodel.train(4000)
\end{verbatim}

\subsection{Inverse linear advection equation code}  \label{A.6}

The numerical results presented in Section \ref{4.6} are obtained from the following code.

\begin{verbatim}
    import src.pinnde as p
    import numpy as np
    import tensorflow as tf

    from scipy.integrate import odeint
    from scipy.fftpack import diff as psdiff
    from scipy.interpolate import RegularGridInterpolator

    c = lambda x: tf.sin(np.pi*x)
    u0 = lambda x: tf.cos(np.pi*x)

    # Creating reference solution
    t = np.linspace(0, 1, 200)
    x = np.linspace(-1, 1, 200)
    c_val = c(x)
    def la(u, t, L):
      ux = psdiff(u, period=L)
      ut = -c_val*ux
      return ut
    def solve_la(u0, t, L):
      sol = odeint(la, u0, t, args=(L,))
      return sol
    ureference = solve_la(u0(x), t, 2)
    uinterp = RegularGridInterpolator((t, x), ureference,
                method='linear')
    
    N_data = 500
    xdata = np.random.uniform(-1, 1, N_data)
    tdata = np.random.uniform(0, 1, N_data)
    data = np.column_stack([tdata, xdata])
    udata = uinterp(data)

    timerange = [0, 1]
    tre = p.domain.Time_NRect(1, [-1], [1], timerange)
    
    bound = p.boundaries.periodic(tre)
    
    u0func = lambda x1: tf.cos(np.pi*x1)
    inits = p.initials.initials(tre, [u0])
    
    dat = p.data.timeinvpinndata(tre, bound, inits, [tdata, xdata],
                                [udata], 1000, 10, 200)
    
    eqn = "ut+c*ux1"
    mymodel = p.models.invpinn(dat, [eqn], ["c"])
    
    rard = p.adaptives.RARD(250, 30, k=3, c=0)
    mymodel.train(5000, adapt_pt=rard)
    
\end{verbatim}

\end{document}